\documentclass[pdflatex,sn-mathphys-num]{sn-jnl}

\usepackage{graphicx}%
\usepackage{multirow}%
\usepackage{amsmath,amssymb,amsfonts}%
\usepackage{amsthm}%
\usepackage{mathrsfs}%
\usepackage[title]{appendix}%
\usepackage{xcolor}%
\usepackage{textcomp}%
\usepackage{manyfoot}%
\usepackage{booktabs}%
\usepackage{algorithm}%
\usepackage{algorithmicx}%
\usepackage{algpseudocode}%
\usepackage{listings}%

\usepackage{multirow} 
\usepackage{bbding}
\usepackage{caption}
\usepackage{graphicx}
\usepackage{booktabs}
\usepackage[accsupp]{axessibility}
\usepackage{hhline}

\theoremstyle{thmstyleone}%
\theoremstyle{thmstyletwo}%

\theoremstyle{thmstylethree}%

\begin{document}

\title[Article Title]{Enhancing Video Transformers for Action Understanding with VLM-aided Training}


\author*[1]{\fnm{Hui} \sur{Lu}}\email{h.lu1@uu.nl}

\author[2]{\fnm{Hu} \sur{Jian}}

\author[1]{\fnm{Albert Ali Salah}}

\author[1]{\fnm{Ronald Poppe}}

\affil*[1]{\orgname{Utrecht University, Utrecht, The Netherlands}}
\affil[2]{\orgname{University of Science and Technology of China, Hefei, China}}



\abstract{Owing to their ability to extract relevant spatio-temporal video embeddings, Vision Transformers (ViTs) are currently the best performing models in video action understanding. However, their generalization over domains or datasets is somewhat limited. In contrast, Visual Language Models (VLMs) have demonstrated exceptional generalization performance, but are currently unable to process videos. Consequently, they cannot extract spatio-temporal patterns that are crucial for action understanding.
In this paper, we propose the Four-tiered Prompts (FTP) framework that takes advantage of the complementary strengths of ViTs and VLMs. We retain ViTs' strong spatio-temporal representation ability but improve the visual encodings to be more comprehensive and general by aligning them with VLM outputs. The FTP framework adds four feature processors that focus on specific aspects of human action in videos: action category, action components, action description, and context information. The VLMs are only employed during training, and inference incurs a minimal computation cost. Our approach consistently yields state-of-the-art performance. For instance, we achieve remarkable top-1 accuracy of 93.8\% on Kinetics-400 and 83.4\% on Something-Something V2, surpassing VideoMAEv2 by 2.8\% and 2.6\%, respectively.

}




\maketitle

\section{Introduction}
\label{sec:intro}
\begin{figure}[t]
  \centering
  \includegraphics[width=1.0\linewidth]{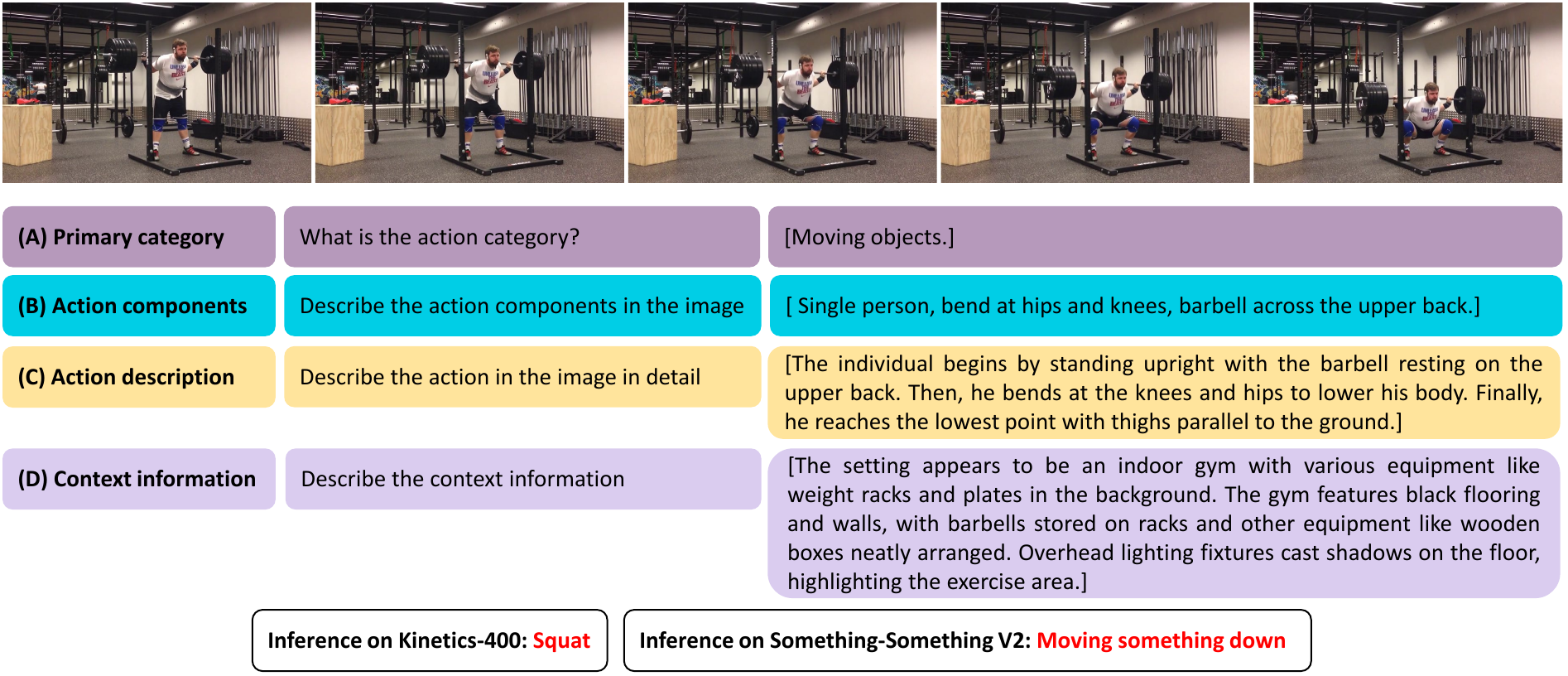}
   \caption{\textbf{Conceptual idea}. Different domains potentially emphasize other aspects of an action's performance, here reflected in the different label sets used in benchmark datasets. VLMs can provide relevant details of video content that might be insufficiently covered by a pretrained visual encoder. In the FTP framework, we employ the textual descriptions from four prompts to align a ViT's visual embeddings during training. This way, we generate richer, comprehensive video representations. During inference, we don't require the VLM and benefit from the more general feature representations to improve action understanding across domains.}
   \label{fig:ftp_concept}
\end{figure}

Video action understanding requires learning rich spatio-temporal representations that reflect human and object appearance and movement, the environment, and interactions between these three aspects. Visual Transformers (ViTs)~\cite{dosovitskiy2021an} are currently the best performing backbones for spatio-temporal representation learning. 
To refine their spatio-temporal representations, researchers have applied different variants of self-attention in the spatial domain of ViTs, including Swin Transformer~\cite{liu2021swin} and local attention~\cite{hu2019local}. These innovations have effectively reduced computational overhead and mitigated redundancy in the spatial domain.
Moreover, various architectures have been proposed to extend Multi-Head Self-Attention (MHSA) along the temporal dimension~\cite{bertasius2021space,arnab2021vivit,yan2022multiview} to capture long-term video dependencies~\cite{li2022uniformerv2}, thus further enhancing the spatio-temporal modeling capabilities of ViTs.

However, when training and benchmarking ViTs for video action understanding, datasets pose different requirements in terms of the relevant spatio-temporal information that needs to be processed. For example, the Kinetics~\cite{carreira2017quo,carreira2019short} datasets focus on action types that are predominantly human-centric, while Something-Something V2 (SSV2)~\cite{goyal2017something} emphasizes the relative movement during interactions. Training on a single dataset potentially leads to poor generalization to another. Ideally, video action understanding models should be able to capitalize on all aspects of the action, including the dynamics, interactions, appearance, and context.

Recently, by combining visual information with Large Language Models, visual language models (VLMs) such as GPT-4~\cite{openai2023gpt4}, BLIP-2~\cite{li2023blip}, and LLaVA~\cite{liu2023visual} have shown improved generalization on various tasks including image captioning~\cite{ming2022visuals} and image understanding~\cite{lu2022learn}. However, current VLMs cannot process videos well due to the shortage of accurately annotated video-text pair, and the computation cost of VLMs during inference is much higher compared to ViTs~\cite{wu2024transferring,zhao2023learning}. This limits their adoption in video action understanding tasks. Recent attempts to extract and concatenate keyframes from videos as the input for VLMs~\cite{himakunthala2023lets} significantly underperform compared to ViTs. One possible reason for this is that temporal information is less readily available from keyframes, especially if these cover a longer time interval. 

We argue that ViTs' strength to extract rich spatio-temporal patterns from videos is partly complementary to VLMs' ability to generalize over various contexts. To take advantage of both, we propose to enhance ViTs' visual encoder with information provided by VLMs. By focusing on aspects that are not directly reflected in action labels, such as scene context, we force the ViT to provide a more comprehensive feature representation that generalizes to different domains and datasets. Since the additional information of VLMs is only used during training, this approach only incurs a minimal computational cost during inference.

To demonstrate the potential of this approach, we introduce the Four-Tiered Prompts (FTP) architecture that relies on the additional information obtained from four VLM prompts, along with the visual encoding obtained from a ViT. Contrastive learning is used to train four feature processors that focus on various aspects of actions in videos: action category, action components, action description, and context information, respectively. Together, these processors provide a rich representation of the action performance. By integrating the outputs of these feature processors, the ViT's generalization ability can be significantly improved. Consequently, we can train ViTs that can flexibly accommodate various label sets, see Figure~\ref{fig:ftp_concept} for an example.
Our main contributions are: 

\begin{enumerate}
    \item We present a novel Four-Tiered Prompts (FTP) framework that focuses on different aspects of the video action. Because the additional VLM inputs are only required during training, the additional computation cost is negligible.
    \item Our work demonstrates the potential of integrating language models into the video domain to enhance model performance.
    \item We report strong performance on video action recognition benchmarks, consistently surpassing state-of-the-art methods by clear margins.
\end{enumerate}

We discuss related works, and then detail our method in Section~\ref{sec:method}. We present our experiments and ablation study in Section~\ref{sec:experiment}, and conclude in Section~\ref{sec:conclusion}.

\section{Related Work}
\label{sec:works}

For video understanding, early works have explored various ways of extracting spatial-temporal information from videos. Video architectures such as 3D ConvNets~\cite{tran2015learning,xie2018rethinking,feichtenhofer2019slowfast} extend 2D image models to the spatio-temporal domain, handling both spatial and temporal dimensions in parallel. There are also 2D ConvNets with temporal modules~\cite{donahue2015long,lin2019tsm,simonyan2014two,wang2018temporal,yue2015beyond}, designed to extend existing 2D CNN models on the temporal dimension. 

\textbf{Vision Transformers}. 
Recent works have adapted vision transformers to the video domain and achieve superior performance compared to previous CNN-based architectures. For example, VTN~\cite{neimark2021video} adopts ViT~\cite{dosovitskiy2021an} to extract spatial features, followed by a Longformer~\cite{beltagy2020longformer} to capture temporal relationships. Both TimeSformer~\cite{bertasius2021space} and ViViT~\cite{arnab2021vivit} further factorize different spatial and temporal attentions for transformer encoders to achieve better performance.
Some works explore how to reduce the computational cost of the space-time attention. MViT~\cite{fan2021multiscale,li2022mvitv2} is a hierarchical transformer with several channel-resolution scale stages, with pooling attention to reduce computation. In contrast, Video
Swin Transformer~\cite{liu2022video} introduces an inductive bias of locality
for videos and achieves a favorable speed-accuracy trade-off. Uniformer~\cite{li2022uniformer} captures local spatio-temporal context and global token dependency by convolution in early layers and by a transformer in deeper layers, respectively. Video MobileFormer~\cite{wang2022video} integrates 3D CNNs and spatial-temporal self-attention mechanisms to enhance efficiency.
Despite their ability to learn spatio-temporal representations from videos, the generalization of current ViTs is not ideal. Specially, when the ViT is trained and performs well on one dataset, its performance significantly potentially deteriorates when tested on dataset that focus on different aspects, such as the use of objects, or the importance of the environment.

\textbf{Visual language models}. Recently, by training on large numbers of image-text pairs, visual language models (VLMs) such as GPT-4~\cite{openai2023gpt4}, BLIP-2~\cite{li2023blip}, and LLaVA~\cite{liu2023visual} have shown excellent generalization performance on computer vision tasks. Current VLMs, however, are currently not capable of processing video data, resulting in subpar performance when dealing with sequential image inputs~\cite{wu2024transferring}. To address this deficit, Lin et al. \cite{lin2023match} extract the most important keyframe and prompt VLMs to recognize the action. Himakunthala et al. \cite{himakunthala2023lets} instead extract multiple keyframes and concatenate them side-by-side as a single image to prompt VLMs. While this approach can reveal various phases of action performance, dynamics cannot be properly extracted. As a result, and despite a significantly higher computation cost during inference, the performance of these methods falls short in comparison to ViTs.

\begin{figure}[t]
  \centering
  \includegraphics[width=1.0\linewidth]{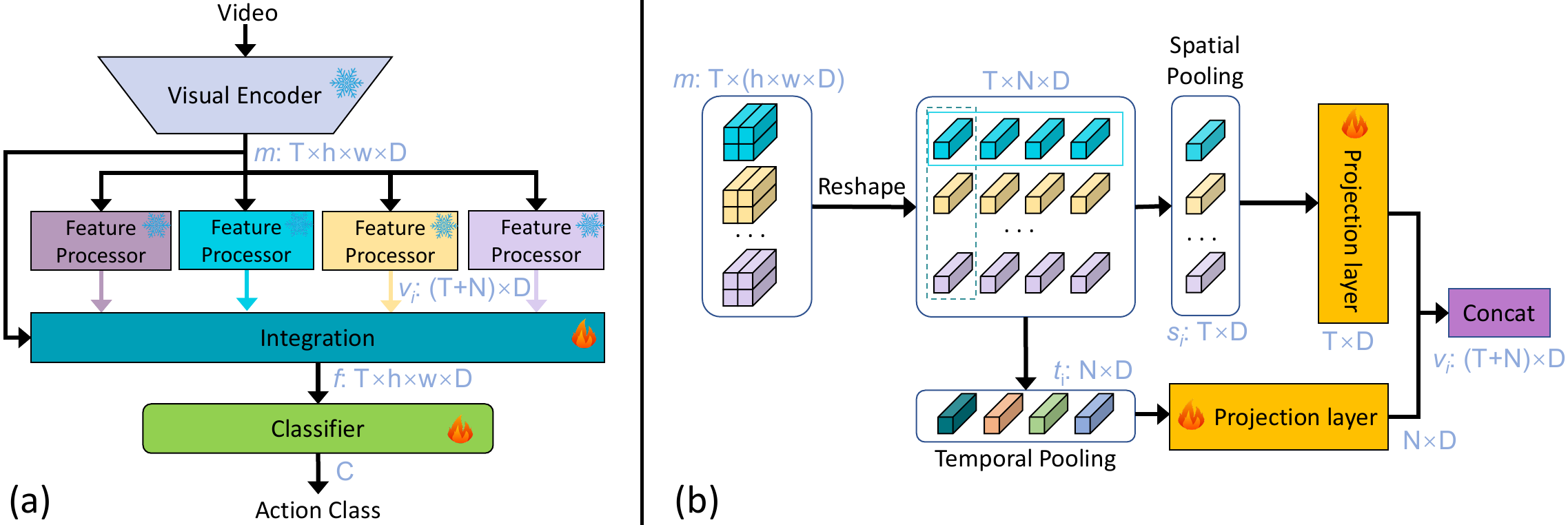}
   \caption{\textbf{Architecture of the FTP framework}. (a) Four feature processors attend to different aspects of the video contents. Their outputs are integrated and subsequently classified. (b) Architecture of a feature processor, with spatial and temporal pooling and projection. The outputs are concatenated to produce the output $v_i$.}
   \label{fig:ftp_architecture}
\end{figure}

\section{Four-Tiered Prompts Framework}
\label{sec:method}
We first provide an overview of our proposed four-tiered prompt (\textbf{FTP}) framework, before discussing the two-stage training process. 

\subsection{Framework}
The FTP architecture is shown in Figure~\ref{fig:ftp_architecture}(a). Taking a ViT as the basis, we add four feature processors after the visual encoder. Each processor is trained to focus on specific aspects of the video contents. The outputs of the four processors are integrated and processed to produce the classification.

\textbf{Visual encoder}. The visual encoder processes the input video and generates feature map $m \in \mathbb{R}^{T \times h \times w \times D}$, with $T$ the number of frames, $h$ and $w$ the height and width of the feature map, and $D$ the depth of the feature map.

\textbf{Feature processor}. We employ four feature processors $i$ ($1 \leq i \leq 4$) to process $m$ and to attend to specific aspects of the video. Refer to Figure~\ref{fig:ftp_architecture}(b) for an overview of the processing steps. In each processor, feature map $m$ is spatially unfolded into a one-dimensional vector of size $N = h \times w$. We then apply pooling along the spatial dimension to produce video-level spatial representation $s_i \in \mathbb{R}^{T \times D}$. In parallel, we apply temporal pooling and obtain a temporal representation $t_i \in \mathbb{R}^{N \times D}$. Both pooled representations are projected onto their input dimensions with a trainable projection layer, and subsequently concatenated to obtain spatio-temporal feature representation $v_i \in \mathbb{R}^{(T+N) \times D}$. The dimensionality of $v_i$ matches that of \cite{xu2021videoclip}, which is crucial for the visual-textual alignment that we perform during training (see Section~\ref{subsec:ftp_training}).

\begin{figure}[htb]
  \centering
  \includegraphics[width=1.0\linewidth]{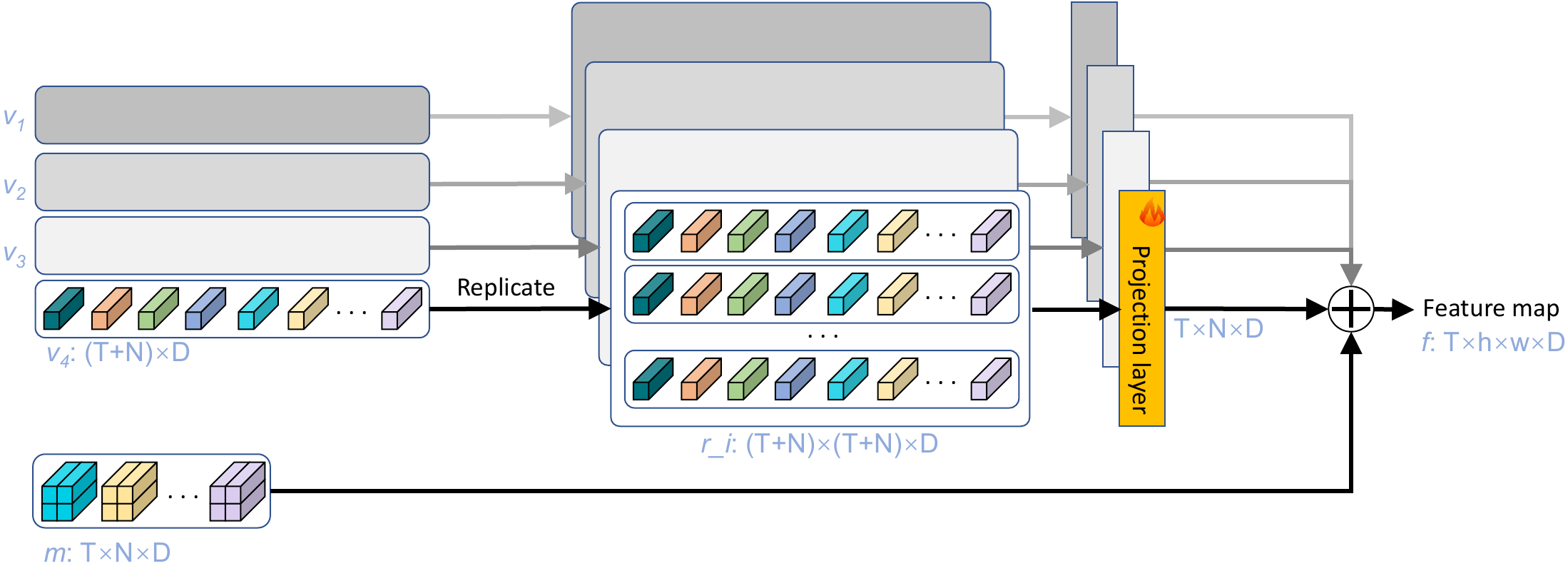}
   \caption{\textbf{Integration process for feature processor and visual encoder outputs}. Each feature processor output $v_i$ is replicated, projected, and finally element-wise summed with visual encoder output $m$.}
   \label{fig:feature_integration}
\end{figure}

\textbf{Integration and classification}. We first integrate the outputs of the feature processors $v_i$ ($1 \leq i \leq 4$) with feature map $m$ from the pretrained visual encoder. To match the dimensions of $v_i$ and $m$, we use an integration layer, shown in Figure~\ref{fig:feature_integration}. We first replicate each feature $v_i$ to produce $r_i$. Subsequently, we use a trainable projection layer to project the video-level feature into the visual encoder’s embedding space. The integrated output of the four feature processors is then element-wise summed with the feature map from the visual encoder, yielding a feature map $f \in \mathbb{R}^{T \times h \times w \times D}$ of the same dimension as $m$.

The final layers consist of two sequential transformer blocks, each containing a multi-head self-attention layer and a feed-forward network layer. After an average pooling layer, the final action classification is performed by a linear layer that maps the activations to the action class outputs. Architecture details appear in the supplementary material~\ref{sm-sec:model architecture}

Compared to the original ViT architecture, the FTP framework only adds computation overhead for the four feature processors. Since these are light-weight, the framework incurs a negligible additional computation cost.

\subsection{Training} \label{subsec:ftp_training}
We need to train the feature processors, their integration, and the classifier. The visual encoder of the ViT remains untouched, which significantly reduces the computational burden of the training. Training proceeds in two stages.

\begin{figure}[htb]
  \centering
  \includegraphics[width=1.0\linewidth]{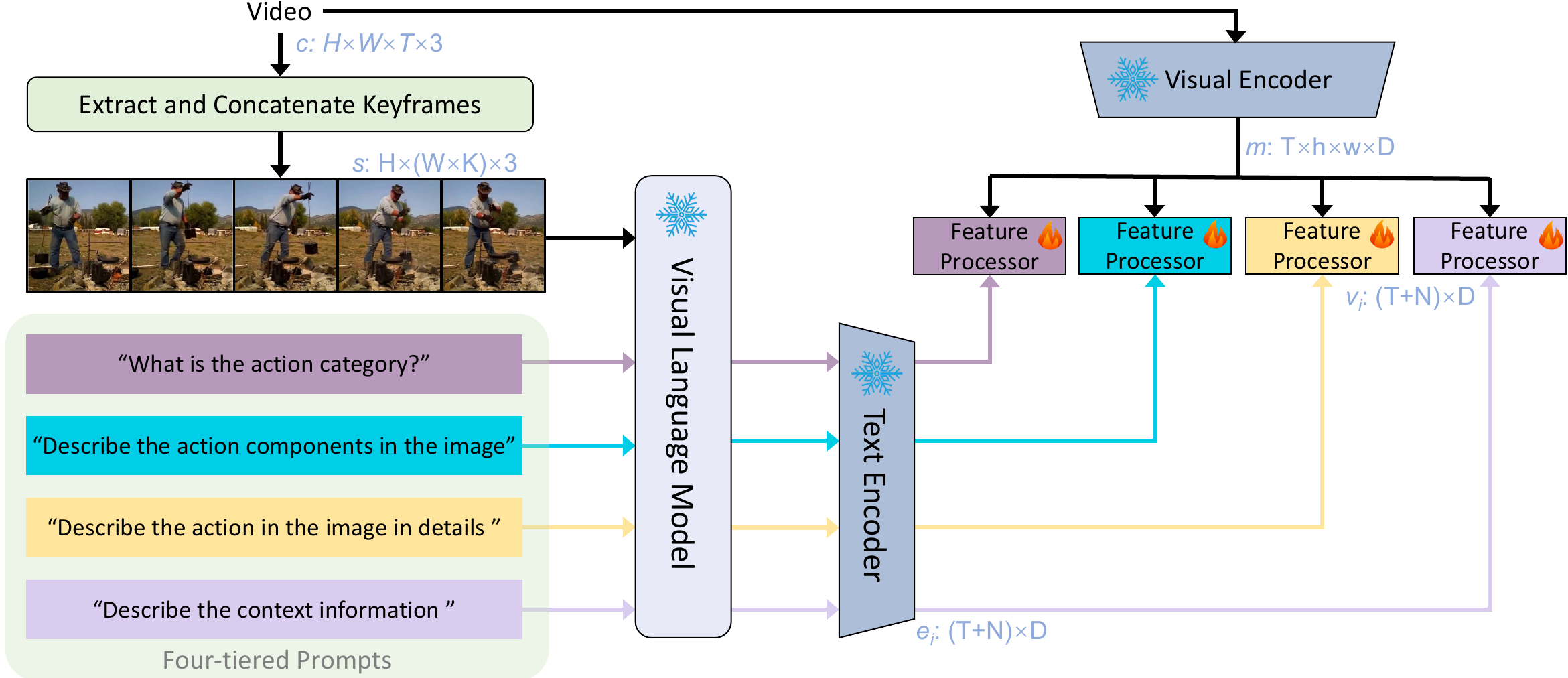}
   \caption{\textbf{Training stage 1}. Concatenated keyframe images are processed by a VLM using four prompts. The outputs pass through a text encoder to yield text embeddings. The feature processors are trained with contrastive loss to project the visual embeddings onto the text embeddings. VLM, text encoder, and visual encoder are frozen.}
   \label{fig:training_stage_1}
\end{figure}

\textbf{Stage 1: Visual-text alignment}. We first train the feature processors using supervision from a VLM and text encoder, shown in Figure~\ref{fig:training_stage_1}. The goal of this stage is to be able to generate rich and general encodings for action videos. This stage only needs to be performed once, to serve model development for a range of target domains. Both VLM and text encoder are used off-the-shelf.

Only during training stage 1, we augment the architecture with an additional processing pipeline that includes the VLM, see Figure~\ref{fig:training_stage_1}. Because VLMs currently cannot process videos directly, we represent a video as a side-by-side concatenation of $K$ keyframes extracted at equal time intervals, similar to \cite{himakunthala2023lets}. The result is a single image $c \in \mathbb{R}^{H \times (W \times K) \times 3}$, with $H$ and $W$ the height and width of the video. We use GPT-4~\cite{openai2023gpt4} (alternatives tested in Section~\ref{sec:ablation}) to generate text descriptions for the keyframe image, according to four specific prompts: (A) ``What is the action category?", (B) ``Describe the action components in the image", (C) ``Describe the action in the image in detail", (D) ``Describe the context information". The four textual outputs are then processed by the pretrained text encoder from~\cite{xu2021videoclip} to produce text embeddings $e_i \in \mathbb{R}^{(T+N) \times D}$ with prompt index $i$ ($1 \leq i \leq 4$).

In parallel, we obtain for the sequence of $T$ frames of the same training video a visual embedding $m$ from the ViT's visual encoder. This embedding passes through the four feature processors $i$ ($i \leq i \leq 4)$, that produce the FTP-based visual encoding $v_i$. We now train the corresponding four feature processors using contrastive learning to transform each visual encoding $v_i$ in such a way that it matches its corresponding textual embeddings $e_i$. Following previous works~\cite{li2021supervision, xu2021videoclip}, we use InfoNCE loss in the training process.

\textbf{Stage 2: Model fine-tuning}.
Based on the availability of rich video encodings, we fine-tune the integration of the four feature processors and subsequent layers for a specific dataset or target domain (Figure~\ref{fig:ftp_architecture}(a)). Apart from a different set of action classes, each domain potentially differs in the importance of different aspects of the human action, such as the dynamics, objects, or the scene. In this stage, we do not require VLM and text encoder, and only train the integration of the feature processors and the classification layers. Owing to the richness of the video encodings that are produced in training stage 1, we learn to prioritize those features that are most meaningful to the target domain.

Specifically, with visual encoder and feature processors frozen, we fine-tune only the projection layer in the feature integration, and the two transformer blocks and final fully-connected layer in the classifier, using cross-entropy loss.

\section{Experiments}
\label{sec:experiment}
We evaluate our FTP framework on several video action recognition benchmarks. We detail the experimental setup in Section~\ref{sec:setup}, present and discuss the main results in Section~\ref{sec:results}, before presenting an ablation study in Section~\ref{sec:ablation}.
\subsection{Experiment setup}
\label{sec:setup}

\textbf{Datasets.} We evaluate our approach on four video action recognition and one action detection datasets: 
(a) \textbf{Kinetics} is among the most widely used action recognition benchmarks. We report on Kinetics-400 (\textbf{K400})~\cite{carreira2017quo} with 400 classes and Kinetics-600 (\textbf{K600})~\cite{carreira2018short} with 600 classes. The sets comprise 240K/20K and 390K/30K training/validation videos with an average duration of 10 seconds, respectively.
(b) \textbf{Something-Something V2 (SSV2)}~\cite{goyal2017something} includes 169K training and 25K validation videos with an average duration of 4 seconds. They are categorized into 174 motion-centric classes of humans performing actions with objects. 
(c) \textbf{UCF-101~\cite{soomro2012ucf101}} is a relatively small dataset, consisting of 9.5K training videos and 3.5K validation videos. 
(d) \textbf{HMDB51~\cite{kuehne2011hmdb}} is also a compact video dataset, containing 3.5K training videos and 1.5K validation videos. We adhere to common evaluation protocols for UCF-101 and HMDB51, averaging results on the three training/validation splits.
(e) \textbf{AVA V2.2}~\cite{gu2018ava} is a challenging dataset for spatio-temporal localization of human actions with 211K training and 57K validation video segments.
Together, these datasets present a significant variation in action performance, label sets, and complexity of the videos, thus allowing for a thorough investigation of the merits of our approach. More experiment results are shown in the supplementary material~\ref{sm-sec:kinetics}.

\begin{table}[t]
  \centering
  \resizebox{\linewidth}{!}{
  \begin{tabular}{llccrrrr}
    \toprule
    Method & Pretrain data &Input & Crops & Params & FLOPs & Top-1 & Top-5 \\
    \midrule
    TimeSformer-L~\cite{bertasius2021space}&IN-21K & $96\times224^2$ & 10$\times$3 & 121M &7.1T & 80.7& 94.7\\
    Mformer-HR~\cite{patrick2021keeping}&IN-21K & $16\times336^2$ &10$\times$3 & 109M &28.8T & 81.1 &95.2\\
    UniFormerV1-B~\cite{li2022uniformer}&IN-1K & $32\times224^2$ &4$\times$3 & 50M &3.1T &83.0& 95.4\\
    Video Swin-L~\cite{liu2022video}&IN-21K & $32\times384^2$ & 10$\times$5 & 200M &105.4T & 84.9& 96.7\\
    ViViT-H~\cite{arnab2021vivit}&JFT-300M & $32\times224^2$   &4$\times$3 & 654M &47.8T & 84.9 &95.8\\
    TokenLearner-L/10~\cite{ryoo2021tokenlearner}&JFT-300M & $64\times224^2$ &4$\times$3 &450M&48.9T & 85.4& 96.3 \\
    MViTv2-L~\cite{li2022mvitv2}&IN-21K & $40\times312^2$ &5$\times$3 & 218M &42.2T &86.1 &97.0\\
    VideoMAE-H~\cite{tong2022videomae}& IN-21K & $16\times224^2$ &5$\times$3 & 633M &17.9T&86.6 &97.1\\ 
    MAE-H~\cite{feichtenhofer2022masked}& IN1K+K400+K600-1.9M & $16\times224^2$ &4$\times$3 & 632M &25.1T &86.8 &97.2 \\
    MaskFeat-L~\cite{wei2022masked}&IN-21K  & $64\times224^2$  &4$\times$3 & 218M &45.5T & 87.0 &97.4\\
    EVL-L/14~\cite{lin2022frozen} &CLIP-400M & $32\times336^2$ &3$\times$3 & 67M &19.1T & 87.7 &--- \\
    X-CLIP-L/14~\cite{ni2022expanding} &CLIP-400M & $16\times336^2$  &4$\times$3 & 453M &37.0T &87.7 &---\\
    MTV-H~\cite{yan2022multiview} &IN-21K+WTS-60M & $32\times224^2$  &4$\times$3 &1120M &44.5T &89.1 &98.2\\
    MTV-H~\cite{yan2022multiview} &IN-21K+WTS-60M & $32\times280^2$ &4$\times$3  & 1330M &73.6T  &89.9 &98.3  \\
    VideoMAE V2-g~\cite{wang2023videomae}&Hybrid-1.35M$^1$ & $64 \times 266^2$ &2$\times$3 & 1050M &160.3T &90.0 &98.4\\ 
    TubeVit-H~\cite{piergiovanni2023rethinking} &IN-1K & $32\times224^2$ &4$\times$3 & 635M &17.6T &90.9 &98.9 \\
    InternVideo~\cite{wang2022internvideo} &CLIP-400M+Hybrid-12M$^2$ & $64 \times 224^2$ &16$\times$4 & 1300M &86.2T & 91.1 &98.9 \\  
    \midrule
    UniFormerV2-B/16~\cite{li2022uniformerv2} &CLIP-400M+K710-0.66M & $8\times224^2$ &4$\times$3 & 115M & 1.8T &85.6 & 97.0 \\
    UniFormerV2-L/14~\cite{li2022uniformerv2} &CLIP-400M+K710-0.66M & $32 \times 224^2$ &2$\times$3 & 354M & 16.0T & 89.3 & 98.2 \\
    UniFormerV2-L/14~\cite{li2022uniformerv2} &CLIP-400M+K710-0.66M  & $64 \times 336^2$ &2$\times$3 & 354M &75.3T& 90.0 &98.4\\
    \midrule
    FTP-UniFormerV2-B/16 (ours)         &CLIP-400M+K710-0.66M & $8\times224^2$  &4$\times$3 & 136M &2.2T &91.9 &98.9 \\
    FTP-UniFormerV2-L/14 (ours)         &CLIP-400M+K710-0.66M & $32\times224^2$ &2$\times$3 & 402M &18.4T & 93.4 & 99.3\\
    \textbf{FTP-UniFormerV2-L/14 (ours)} &CLIP-400M+K710-0.66M & $64\times336^2$ &2$\times$3 & 402M & 78.7T &\textbf{94.3} &\textbf{99.4}\\
    \bottomrule
  \end{tabular}
  }
  \caption{Performance on K400. ---: numbers not reported. Best results in \textbf{bold}. We report crops (temporal $\times$ spatial) and TFLOPs for inference. \\$^1$ K710+SSV2+WebVid2M+AVA V2.2, and videos crawled from Instagram\\ $^2$ WebVid2M/10M+HowTo100M+K710+SSV2+AVA V2.2, and self-collected videos.}
  \label{tab:kinetics400}
\end{table}

\textbf{Implementation details}. Our FTP framework can be implemented with a range of ViTs. For our main results, we use the well-performing visual encoder from UniformerV2~\cite{li2022uniformerv2}. It is based on CLIP-ViT~\cite{radford2021learning}, with different capacities (ViT-B, ViT-L), and pretrained on CLIP-400M. We insert the global UniBlocks in the last four layers to perform the multi-stage fusion. For our resulting FTP-UniFormerV2-B/16 and FTP-UniFormerV2-L/14, we use $T=16$ and $T=32$ franes, respectively. We use a resolution of 224$^2$, and additionally report on K400 with a resolution of 336$^2$ and $T=64$. The number of parameters of these models ranges from 136--402, with TFLOPs between 2.2--78.7.

Following \cite{li2022uniformerv2,wang2022internvideo}, when reporting on K400/K600, we train on Kinetics-710 (K710). K710 combines the Kinetics-400/600/700 datasets, while removing test videos. For all other datasets, we fine-tune the pretrained models on the training set and report on the validation set. For action detection, we additionally predict bounding boxes before classification. We use Faster-RNN as in \cite{li2022uniformerv2}.

In training stage 1, we set the base learning rate to $1.0 \times 10^{-5}$, repeated sampling to 1, batch size to 512, and train for 800 epochs. We adopt sparse sampling~\cite{wang2016temporal}. The number of keyframes is set to $K = 5$. For training stage 2, we keep these settings but set the learning rate to $1.5 \times 10^{-6}$ and train for 40 epochs. More details appear in the supplementary material~\ref{implementation}. All experiments are performed on 16 NVIDIA A100 GPUs.

\begin{table}[t]
  \centering
  \resizebox{\linewidth}{!}{
  \begin{tabular}{llcccccc}
    \toprule
    Method & Input &Crops &Params (M) & TFLOPs &Top-1 & Top-5\\
    \midrule
    TimeSformer-HR~\cite{bertasius2021space} &($16 \times 448^2$) &16×3 &121 & 5.1 &62.4 &86.3 \\
    SlowFast R101~\cite{feichtenhofer2019slowfast}&($64 \times 224^2$)  &8×8 &53 &0.3 &63.1 &87.6\\
    ViViT-L FE~\cite{arnab2021vivit}  &($32 \times 224^2$)  &1×3 &612 & 47.6 &65.4 &89.8\\
    EVL-L/14~\cite{lin2022frozen}&($32 \times 336^2$) &3×3 & 67 &9.6 &66.7 &---\\
    Mformer-L~\cite{patrick2021keeping}&($32 \times 224^2$)  &1×3 &109 & 3.6 &68.1 &91.2\\
    VIMPAC~\cite{tan2021vimpac}&($10 \times 256^2$)  &10×3 &307 &--- &68.1 &---\\
    TDN R101~\cite{wang2021tdn}&($24 \times 224^2$)  &10×3 &88 &0.6 &68.2& 91.6\\
    MTV-B~\cite{yan2022multiview}& ($32 \times 320^2$) &4×3 &310 &11.2 &68.5& 90.4 \\
    Video Swin-B~\cite{liu2022video}&($32 \times 224^2$)  &1×3 &89 & 1.0 &69.6 &92.7 \\
    UniFormerV1-B~\cite{li2022uniformer}&($32 \times 224^2$)  &1×3& 50 &0.8 &71.2 &92.8\\
    BEVT~\cite{wang2022bevt}&($32 \times 224^2$)  &--- &88 &1.0 & 71.4 & ---\\ 
    MViTv2-B~\cite{li2022mvitv2}&($32 \times 224^2$)  &1×3& 51&0.7 &72.1 &93.4 \\
    MaskFeat-L~\cite{wei2022masked} &($64 \times 312^2)$ &4×3 &218 &8.5 &75.0 &95.0 \\
    VideoMAE-L~\cite{tong2022videomae} &($32 \times 224^2$)  &1×3 &305 & 4.3  &75.4 &95.2\\
    TubeViT-L~\cite{piergiovanni2023rethinking} &($32 \times 224^2$)  &4×3 &311 & 9.5 &76.1 &95.2 \\
    VideoMAE V2-g~\cite{wang2023videomae} &($64 \times 266^2$) &5×3 &1050 &160.3 & 77.0 &95.9\\
    InternVideo~\cite{wang2022internvideo} &($64 \times 224^2$) &16×4 & 1300 &86.2 &77.2 &95.9 \\
    \midrule
    UniFormerV2-B/16~\cite{li2022uniformerv2} &($16\times224^2$) &1$\times$3 & 163 & 0.6 &69.5 & 92.3 \\
    UniFormerV2-L/14~\cite{li2022uniformerv2} &($32 \times 224^2$)  &1×3& 574 & 5.2 &73.0 &94.5 \\
    \midrule
    FTP-UniFormerV2-B/16 (ours) &($16 \times 224^2$)  &1×3 &183 &1.1 &77.3 &96.7 \\
    \textbf{FTP-UniFormerV2-L/14 (ours)} &($32 \times 224^2$)  &1×3 &620 &6.6 &\textbf{79.8} &\textbf{98.9} \\
    \bottomrule
  \end{tabular}
  }
  \caption{Performance on SSV2. ---: numbers not reported. Best results in \textbf{bold}.}
  \label{tab:ssv2}
\end{table}

\subsection{Main results}
\label{sec:results}
We first address action recognition, before moving to action detection.

\textbf{Kinetics-400}. Results on K400 are summarized in Table~\ref{tab:kinetics400}. Our FTP models outperform previous top-perfoming methods by a clear margin. The best performing FTP-UniFormerV2-L/14 with $64 \times 336^2$ inputs achieves state-of-the-art performance with an unprecedented 94.3\%. But even the significantly smaller FTP-UniFormerV2-B outperforms the previous best InternVideo~\cite{wang2022internvideo}, despite having only 10.5\% of the parameters, $T=8$ instead of 64 input frames, and having seen a significantly lower number of training videos. Compared to their vanilla counterparts, FTP-UniFormerV2 models achieve 4.3-6.1\% higher top-1 on the already saturated results.

\textbf{Something-Something V2.} Table~\ref{tab:ssv2} presents comparisons to the state-of-the-art methods on SSV2. Our smaller FTP-UniFormerV2-B/16 model performs on par with the previous best approach InternVideo~\cite{wang2022internvideo} while having only ~1.3\% of the TFLOPs and 14.1\% of the parameters. Our increased performance thus is not due to additional model capacity or computation cost. Our FTP-UniFormerV2-L/14 model achieves state-of-the-art performance of 79.8\%, outperforming InternVideo and VideoMAE V2-g~\cite{wang2023videomae} by 2.6--2.8\%.

\textbf{UCF-101 and HMDB51.} From Table~\ref{tab:UCF101-HMDB51}, for UCF-101, we observe that our FTP-UniFormerV2-L/14 performs on par with the top model VideoMAE V2~\cite{wang2023videomae}, despite having much less parameters. For HMDB51, our method outperforms all tested methods, including the previous best VideoMAE V2~\cite{wang2023videomae} (+3.4\%) while requiring only 38.3\% of the parameters.

\begin{table}[ht]
  \centering
  \begin{minipage}[t]{0.5\linewidth}
    \centering
    \resizebox{\linewidth}{!}{
  \begin{tabular}{lccc}
    \toprule
    Method & Params & UCF-101 &HMDB51 \\
    \midrule
    VideoMoCo~\cite{pan2021videomoco} & 15M &78.7  &49.2\\
    MemDPC~\cite{han2020memory} & 32M &86.1   &54.5\\
    Vi$^2$CLR~\cite{diba2021vi2clr} & 9M &89.1 &55.7\\
    VIMPAC~\cite{tan2021vimpac} & 307M& 92.7& 65.9\\
    CORP~\cite{hu2021contrast} & 32M& 93.5 &68.0\\
    XDC~\cite{alwassel2020self} &33M &94.2 &67.1 \\ 
    CVRL~\cite{qian2021spatiotemporal} &  328M& 94.4& 70.6\\
    GDT~\cite{patrick2020multimodal} &33M &95.2 &72.8 \\
    MMVFAC~\cite{alayrac2020self}  &94M &95.2 &75.0\\
    VideoMAE~\cite{tong2022videomae} & 87M& 96.1 &73.3\\
    MVD-L~\cite{wang2023masked} &87M&  97.5 &79.7\\
    VideoMAE V2~\cite{wang2023videomae}&1050M & 99.6 &88.1\\
     \midrule
    UniFormerV2-B/16~\cite{li2022uniformerv2} & 115M &96.8 & 80.0 \\
    UniFormerV2-L/14~\cite{li2022uniformerv2} & 354M &98.2& 86.2 \\
    \midrule
    FTP-UniFormerV2-B/16 (ours)&136M &98.7 &83.9 \\
    \textbf{FTP-UniFormerV2-L/14 (ours)}&402M&\textbf{99.7} &\textbf{91.5} \\
    \bottomrule
  \end{tabular}
    }
    \caption{Results on UCF-101/HMDB51.}
    \label{tab:UCF101-HMDB51}
  \end{minipage}
  \hfill
  \begin{minipage}[t]{0.45\linewidth}
    \centering
    \resizebox{\linewidth}{!}{
    \begin{tabular}{lcccc}
    \toprule
    Method & FLOPs &Param& mAP  \\
    \midrule
    SlowFast R101~\cite{feichtenhofer2019slowfast}   &138G& 53M &23.8\\
    HIT~\cite{faure2023holistic}&622G & 198M&32.6\\
    MViTv2-L~\cite{li2022mvitv2} & 2828G &213M &34.4\\
    ST-MAE-H~\cite{feichtenhofer2022masked}   &1193G& 632M &36.2\\
    VideoMAE-L~\cite{tong2022videomae}  &597G &305M &37.0\\
    MaskFeat~\cite{wei2022masked} &2828G &218M &37.5 \\
    VideoMAE-H~\cite{tong2022videomae}  &1192G& 633M &39.5\\
    InternVideo~\cite{wang2022internvideo}  &8733G& 1300M &41.0\\
    MVD-H~\cite{wang2023masked}  & 1192G &633M &41.1\\
    VideoMAE V2~\cite{wang2023videomae}   &4220G& 1050M &42.6\\
    LART-MViT~\cite{rajasegaran2023benefits} &3780G & 1640M&42.6\\
    Hiera-H~\cite{ryali2023hiera}&1158G &672M &43.3\\
     \midrule
    UniFormerV2-B/16~\cite{li2022uniformerv2} & 406G &115M&38.4 \\
    UniFormerV2-L/14~\cite{li2022uniformerv2} & 833G &354M&40.3 \\
    \midrule
    FTP-UniFormerV2-B/16 (ours) &482G &136M &43.9 \\
   \textbf{FTP-UniFormerV2-L/14 (ours)} &970G  &402M &\textbf{46.2}\\
    \bottomrule
  \end{tabular}
  }
    \caption{Results on AVA V2.2.}
    \label{tab:AVA}
  \end{minipage}
\end{table}

\textbf{AVA V2.2.} Also with an additional challenge of localizing the action, our approach shows remarkable improvement over previous methods, shown in Table~\ref{tab:AVA}. While our smaller FTP-UniFormerV2-B/16 already performs better than the previous best Hiera-H~\cite{ryali2023hiera} (+0.6\%), the larger FTP-UniFormerV2-L/14 achieves a 2.9\% performance gain. Our performance is currently limited due to the presence of multiple targets in a sequence, because our visual encoder is trained on single targets. This complicates the alignment of the feature processor with the text descriptions generated by the VLM.

\subsection{Ablation study}
\label{sec:ablation}
We have observed consistent improvements when using FTP. In this section, we investigate the role of the VLM, visual encoder, and the prompts. We also provide qualitative analyses. Additional results appear in the supplementary material~\ref{sm-sec:kinetics}.

\begin{table}[t]
  \centering
  \resizebox{\linewidth}{!}{
  \begin{tabular}{l|cc|cc|cc|ccc}
    \toprule
    \multirow{2}{*}{VLM} & \multicolumn{2}{c|}{K400} & \multicolumn{2}{c|}{K600} & \multicolumn{2}{c|}{SSV2} &UCF-101 & HMDB51 & AVA V2.2 \\
    &Top-1 &Top-5 &Top-1 &Top-5  &Top-1 &Top-5 &Top-1 &Top-1 &mAP\\
    \midrule
    OpenFlamingo~\cite{awadalla2023openflamingo}& 91.2 &98.9 &91.3 &98.9 &70.2 &91.8 &98.4 &88.5 &41.9\\
    BLIP-2~\cite{li2023blip}                & 92.8 &99.0 &91.7 &98.9 &77.1 &95.4 &99.1 &90.7 &43.6\\
    MiniGPT-4~\cite{zhu2023minigpt}         & 93.1 &99.2 &93.2 &99.1 &73.3 &94.1 &99.1 &89.7 &43.2\\
    MiniGPT-v2~\cite{chen2023minigpt}       & 93.3 &\textbf{99.3} &93.6 &99.2 &73.6 &94.0 &99.6 &\textbf{91.5} &43.6\\
    LLaVA~\cite{liu2023visual}              & \textbf{93.4} &\textbf{99.3} &93.4 &99.2 &79.6 &98.6 &99.4 &90.3 &44.7\\
    GPT-4~\cite{openai2023gpt4}             & \textbf{93.4} &\textbf{99.3} &\textbf{93.8} &\textbf{99.4} &\textbf{79.8} &\textbf{98.9} &\textbf{99.7} &\textbf{91.5} &\textbf{46.2}\\
    \bottomrule
  \end{tabular}
  }
  \caption{Results on FTP-UniFormerV2-L/14 for different VLMs. Best results in \textbf{bold}.}
  \label{tab:different LVLMs}
\end{table}

\textbf{Effect of VLM}. Various VLMs can be used to generate text descriptions for keyframe images. We experiment with text representations generated by OpenFlamingo~\cite{awadalla2023openflamingo}, MiniGPT-4~\cite{zhu2023minigpt}, MiniGPT-v2~\cite{chen2023minigpt}, LLaVA~\cite{liu2023visual}, BLIP-2~\cite{li2023blip}, and GPT-4~\cite{openai2023gpt4}. For GPT-4, we directly use their web API. For all other models, we employ their publicly available code and model weights. For training, we use the FTP-UniFormerV2-L/14 with $32 \times 224^2$ inputs, with the same settings as before and calculate the model accuracy on various video action datasets.

The results are shown in Table~\ref{tab:different LVLMs}. At a first glance, the performance of OpenFlamingo is somewhat lower. Compared to Flamingo, an overall relative performance of 80-89\% is reported \cite{awadalla2023openflamingo}, which shows to be insufficient to compete with the other VLMs. GPT-4 shows the best performance, likely because its more comprehensive scene descriptions, and its more extensive training.

The difference between the top five VLMs on K400 is only 0.6\%. We hypothesize that the similar performance is mainly due to the human-centric nature of the Kinetics videos. Object and scene information might be less relevant for the action labels. In turn, differences in the quality and richness of the VLM outputs might have had only a limited effect. We also see relatively similar performances across VLMs for human-centric datasets K600, UCF-101, and HMDB51.

For SSV2, we obtain similar performances for GPT-4 (79.8) and LLaVA (79.6\%), while MiniGPT-4 (73.6\%) and MiniGPT-v2 (73.3\%) show poorer performance than BLIP-2 (77.1). The VLM's ability to understand relative movement is crucial to classify SSV2 videos. We conclude that some VLMs are less able to extract such patterns from the concatenated keyframe images.

For action detection results on AVA V2.2, the difference between VLMs is also significant. Compared to GPT-4 (46.2\%), other VLMs show lower performance, somewhat modest for LLaVA (-1.5\%) but larger for MiniGPT-v2 (-2.6\%), BLIP-2 (-2.6\%), MiniGPT-4 (-3.0\%), and OpenFlamingo (-4.3\%). We suspect that this is because the AVA V2.2 dataset involves multiple targets and the action types present in the clips are more diverse. Consequently, a more fine-grained content description is required. Differences in how comprehensive a textual output is, has a larger effect on the final classification performance.

\begin{table}[t]
\centering
\resizebox{\linewidth}{!}{
\begin{tabular}{l|ll|ll|ll|l}
\toprule
\multirow{2}{*}{ViT} & \multicolumn{2}{c|}{K400}        & \multicolumn{2}{c|}{K600}        & \multicolumn{2}{c|}{SSV2}        & AVA V2.2\\ 
                         & \multicolumn{1}{l}{Top-1} & Top-5 & \multicolumn{1}{l}{Top-1} & Top-5 & \multicolumn{1}{l}{Top-1} & Top-5 & mAP \\ \midrule
ViT~\cite{dosovitskiy2021an}  & \multicolumn{1}{l|}{81.6}     &  95.1  & \multicolumn{1}{l|}{83.5}     &  96.2    & \multicolumn{1}{l|}{67.5}     & 91.0     &  28.7 \\ 
ViT + FTP                        & \multicolumn{1}{l|}{84.8 (+3.2)}     &  95.8  & \multicolumn{1}{l|}{85.7 (+2.2)}     & 96.5     & \multicolumn{1}{l|}{68.9 (+1.4)}     &   91.4   &  30.0 (+1.3)   \\ \midrule

DeiT III~\cite{touvron2022deit}    & \multicolumn{1}{l|}{82.7}     &  95.5    & \multicolumn{1}{l|}{84.0}     &  96.5    & \multicolumn{1}{l|}{69.6} & 92.7  &  29.0  \\ 
DeiT III~\ + FTP                   & \multicolumn{1}{l|}{85.7 (+3.0)}   &97.0 & \multicolumn{1}{l|}{86.2 (+2.2)} & 97.3     & \multicolumn{1}{l|}{70.6 (+1.0)} & 92.0   & 30.8 (+1.8)  \\ \midrule
VideoMAE-L~\cite{tong2022videomae} & \multicolumn{1}{l|}{85.2}     & 96.8     & \multicolumn{1}{l|}{86.1} &  97.0  & \multicolumn{1}{l|}{75.4}     &  95.2    &   37.0  \\ 
VideoMAE-L + FTP           & \multicolumn{1}{l|}{88.3 (+3.1)} & 97.6  & \multicolumn{1}{l|}{89.0 (+2.9)}   & 98.2   & \multicolumn{1}{l|}{77.0 (+1.6)}  &95.9    &39.5 (+2.5)    \\ \midrule
X-CLIP-L~\cite{ni2022expanding}& \multicolumn{1}{l|}{87.7}     &  97.4    & \multicolumn{1}{l|}{88.3}     &  97.7  & \multicolumn{1}{l|}{72.2}  &  93.7 &  36.1   \\ 
X-CLIP-L + FTP           & \multicolumn{1}{l|}{89.6 (+1.9)} &  98.3    & \multicolumn{1}{l|}{90.1 (+1.8)} & 98.3     & \multicolumn{1}{l|}{73.1 (+0.9)}   & 94.5     & 38.2 (+2.1)    \\ \midrule
ILA~\cite{tu2023implicit}                & \multicolumn{1}{l|}{88.7}     & 97.8     & \multicolumn{1}{l|}{89.2} &  98.2  & \multicolumn{1}{l|}{70.2}   &  91.8    &   36.0  \\
ILA + FTP               & \multicolumn{1}{l|}{91.0 (+2.3)} &  98.7    & \multicolumn{1}{l|}{90.6 (+1.4)} & 98.5     & \multicolumn{1}{l|}{71.5 (+1.3)}     &  93.5    & 37.9 (+1.9)    \\ \midrule
MTV-H~\cite{yan2022multiview} & \multicolumn{1}{l|}{89.1}     &  98.2    & \multicolumn{1}{l|}{89.6}     &  98.3    & \multicolumn{1}{l|}{67.6}   &  90.1    &  31.2   \\ 
MTV-H + FTP                    & \multicolumn{1}{l|}{91.4 (+2.3)}     &   98.7  & \multicolumn{1}{l|}{90.9 (+1.3)}     & 98.5   & \multicolumn{1}{l|}{69.1 (+1.5)}     &  92.6    &  33.2 (+2.0)   \\  \midrule

UniformerV2-L~\cite{li2022uniformerv2}  & \multicolumn{1}{l|}{89.3} & 98.2  & \multicolumn{1}{l|}{89.5}     & 98.3    & \multicolumn{1}{l|}{73.0} &94.5   & 40.3 \\ 
UniformerV2-L + FTP & \multicolumn{1}{l|}{\textbf{93.4} (+4.1)} & \textbf{99.3}  & \multicolumn{1}{l|}{\textbf{93.8} (+4.3)}  & \textbf{99.4}    & \multicolumn{1}{l|}{\textbf{79.8} (+6.8)}  &\textbf{98.9}    & \textbf{46.2} (+5.9) \\ \bottomrule
\end{tabular}
}
\caption{Results for different ViTs with and without FTP. Best results in \textbf{bold}.}
\label{tab:generality}
\end{table}

\textbf{Effect of visual encoder}. We apply our approach with visual encodings from different ViTs: vanilla ViT~\cite{dosovitskiy2021an}, DeiT III~\cite{touvron2022deit}, VideoMAE~\cite{tong2022videomae}, X-CLIP~\cite{ni2022expanding}, ILA~\cite{tu2023implicit}, and MTV-H~\cite{yan2022multiview}. We again use inputs of $32 \times 224^2$ and  keep all other settings. In Table~\ref{tab:generality}, we summarize the results for the various ViTs with and without our FTP framework.

The use of FTP consistently improves the performance across datasets and ViTs. We obtain performance gains in the range of 0.9--6.8\%. The largest gains are for UniFormerV2, which already provided the best results across the tested ViTs for K400, K600, and AVA V2.2. Without FTP, VideoMAE-L produced the best results on SSV2. This could be attributed to the limited generality of the original UniformerV2-L, which was solely trained on K710. When the FTP framework is used to improve the visual encodings, the performance gain is significant at 6.8\%, and makes FTP-UniFormerV2-L the top performing model.

\begin{table}[t]
\centering
\resizebox{\linewidth}{!}{
\begin{tabular}{c|cccc|cc|cc|cc|ccc}
\toprule
\multirow{2}{*}{Group} & \multicolumn{4}{c|}{Prompts}& \multicolumn{2}{c|}{K400} & \multicolumn{2}{c|}{K600}& \multicolumn{2}{c|}{SSV2}     & UCF-101 & HMDB51   & AVA V2.2 \\ 
                        & A                    & B                    & C                    & D                    & Top-1 & Top-5 & Top-1 & Top-5 & Top-1 & Top-5 & Top-1 & Top-1 & mAP    \\ \midrule
1                  &   &   &   &   &87.5 &97.6&89.0 & 98.0  &66.7&91.3 & 97.8  &80.5  &39.5  \\ \midrule
2                  & \Checkmark   &   &   &   &88.6 & 98.2 & 89.9  & 98.2 &69.4&92.1  & 98.2  & 83.8 &41.1        \\
3                  &   & \Checkmark   &   &   &89.0 & 98.4 & 90.5  & 98.8 &71.2&92.8  & 98.6  & 86.0 & 42.2       \\
4                  &   &   & \Checkmark   &   &88.8 & 98.2 & 89.4  & 98.0 &72.0&93.9  & 98.4  & 86.8 & 42.6      \\
5                  &   &   &   & \Checkmark   &88.5 & 98.2 & 90.0  & 98.2 &71.0&92.8  & 98.2  & 85.6 & 42.0       \\ \midrule
6                  & \Checkmark   & \Checkmark   &   &   &91.3 & 98.9 & 91.7  & 99.0 &72.3&93.9  & 99.0  & 87.2 & 42.8       \\
7                  & \Checkmark   &   & \Checkmark   &   &90.9 & 98.9 & 91.3  &99.0  &73.3&94.2  & 99.0  & 88.3 & 43.4      \\
8                  & \Checkmark   &   &   & \Checkmark   &90.2 & 98.6 & 90.7  & 98.8 &71.9&93.8  & 98.9  & 86.8 & 42.6       \\ 
9                  &   & \Checkmark   & \Checkmark   &   &92.3 & 99.0 & 92.7  & 99.1 &75.3&95.2  & 99.3  & 89.5 & 44.6       \\ 
10                 &   & \Checkmark   &   & \Checkmark   &91.5 & 98.9 & 91.9  & 99.0 &74.7&94.8  & 99.2  & 89.1 &44.2     \\ 
11                 &   &   & \Checkmark   & \Checkmark   &91.9 & 98.9 & 92.3  &99.1  &75.0&95.0  & 99.2  &89.2  & 44.4       \\ \midrule
13                 & \Checkmark   & \Checkmark   & \Checkmark   &   &92.5 & 99.0 & 92.9 & 99.3 &78.9&98.5 &  99.4 & 90.7 & 45.8       \\ 
14                 & \Checkmark   & \Checkmark   &   & \Checkmark   &93.2 & \textbf{99.3} & 93.3  & 99.3 &77.8&97.3 & 99.6  & 89.7  & 44.9       \\ 
15                 & \Checkmark   &   & \Checkmark   & \Checkmark   &92.9 & 99.1 &93.0   & 99.3 &78.0&97.4 & 99.5  & 90.3 & 45.2      \\ 
16                 &     & \Checkmark   & \Checkmark & \Checkmark   &92.6 & 99.1 & 92.9  & 99.3 &79.1&98.6 &  99.4 & 90.8 & 45.8       \\ \midrule
17                 & \Checkmark   & \Checkmark   & \Checkmark   & \Checkmark   &\textbf{93.4} & \textbf{99.3} & \textbf{93.8} &\textbf{99.4}  &\textbf{79.8}&\textbf{98.9} & \textbf{99.7}  & \textbf{91.5} & \textbf{46.2}    \\ \bottomrule
\end{tabular}
}
\caption{Results on FTP-UniFormerV2-L/14 for different combinations of prompts.}
\label{tab:prompts function}
\end{table}

\textbf{Effect of prompt types}. In the FTP framework, we use four feature processors that each attend to a specific aspect of the video contents. In training stage 1, they are aligned with text encodings of four corresponding prompts (refer to Section~\ref{sec:method} and Figure~\ref{fig:ftp_concept}). While the prompts have been chosen to cover various perspectives, we have not investigated their relative contribution until now. For these experiments with different combinations of the four prompts, we use the FTP-UniFormerV2-L/14 with $32 \times 224^2$ input and keep all other settings as before. We  report the models' accuracy on various video action datasets in Table~\ref{tab:prompts function}. We refer to the various combinations with their group number.

Group 1 only relies on the visual encoder, without additional inputs coming from the VLM. Because of architectural differences in the classification layers between our FTP-UniFormerV2-L/14 and the vanilla UniFormerV2-L/14 from ~\cite{li2022uniformerv2}, the performance of the two models differs.

Overall, we see a trend of increasing performance when we use more prompts. From zero to four prompts, there is a 5.9\% and 13.1\% increase for K400 and SSV2, respectively. We observe a similar increasing trend for other datasets. When using a single prompt, we observe the largest increase for prompt B on K400, K600, and UCF-101, while SSV2, HMDB51 and AVA V2.2 benefit most from prompt C. This fact seems to suggest that different datasets focus on different characteristics of the action and its depiction. When two prompts are used in groups 6--11, we see superior performance when prompts B and C are used together.

With three prompts, we notice that leaving out either prompt B or C produces the best results for K400, K600, and UCF-101, but the lowest for the other tested datasets. We thus see the same division as before, again confirming our hypothesis that different datasets concentrate on different aspect of the action.

Finally, when all four prompts are used, we obtain the best results. Overall, each prompt provides complementary information, with a combination of all four prompts to consistently yield the best results. This analysis suggests that adding more prompts is generally beneficial. Future work should be aimed at understanding which aspects of the action performance or depiction we are currently not captured, and how to include the missing information. The FTP framework for combining including VLM textual outputs into the visual encodings coming from the ViT, presented in this paper, provides an effective tool to conduct such investigations.

\begin{figure}[htb]
  \centering
  \includegraphics[width=1.0\linewidth]{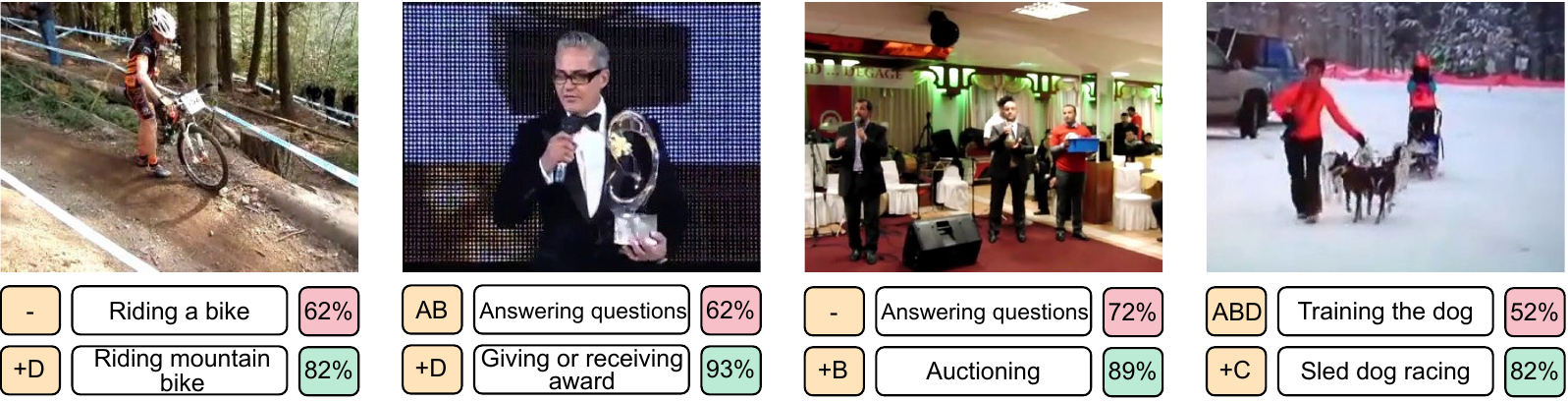}
   \caption{Example classifications for two different combinations of prompts, with softmax outputs. Top row is misclassified, bottom row is correctly classified.}
   \label{fig:qualitative_evaluation}
\end{figure}

\textbf{Qualitative analysis}. We show several frames of example test videos from K400 that are classified differently when using another combination of prompts. The VLM is not used during inference so we cannot analyze the textual descriptions for each prompt. Still, we can try to understand how additional prompts may have helped to fix misclassifications.

For example, the leftmost two examples in Figure~\ref{fig:qualitative_evaluation} benefit from the additional input from the feature processor that has been aligned to encode context information. In the third example, the additional availability of the action components could reveal the movements of all people. The rightmost example could benefit from additional granularity encoded by the third feature processor. The supplementary material~\ref{sm-sec:kinetics} contains more quantitative and qualitative analyses.

\section{Conclusion and Future Work}
\label{sec:conclusion}
We have introduced Four-Tiered Prompts (FTP), a framework to include semantic information into the visual encodings of Video Transformers (ViTs). During training only, we leverage a visual language model (VLM) to produce textual descriptions of a video, with four prompts that focus on different aspects of the video content. We train feature processors to align the output of the visual encoder with these text descriptions, to produce more comprehensive visual embeddings. Owing to these richer representations, we then fine-tune for action understanding in a broad range of domains, and with different requirements such as an emphasis on objects or dynamics. The feature processors incur a minimal computational overhead during inference. Moreover, VLM, text encoder, and visual encoder are used off-the-shelf, which makes the approach flexible.

Experiments on Kinetics-400/600, Something-Something V2, HMDB51, UCF-101, and AVA V2.2 consistently demonstrate state-of-the-art performance, while the computation cost is typically lower than recent methods. We have further validated the influence of the VLM, ViT, and the combination of prompts.

There are several limitations and avenues for future work. First, the additional feature processors operate on the output of the ViT's visual encoder, which typically has a limited dimensionality and potentially overlooks important aspects of the video contents. We therefore expect that integration of the textual embeddings at an earlier stage is beneficial for the performance. Second, our FTP framework employs four prompts that focus on different aspects of action performance but a more systematic analysis of the number of type of prompts could reveal which aspects are not sufficiently covered. We argue that such investigations will further improve the performance of the FTP framework. Together with its flexibility, we expect that the framework will be effective in a range of application domain, potentially also beyond action understanding.

\bibliography{sn-bibliography}

\newpage
\begin{appendices}
\begin{table}[htbp]
  \centering
  \resizebox{\linewidth}{!}{
  \begin{tabular}{cc}
    \toprule
    Stage & FTP-UniFormerV2-L \\
    \midrule
    Visual encoder & [3,4,8] Uni.V2 Blocks\\
    \hline
    \multirow{5}{*}{Feature processor} &  torch.transpose(features, 1, -1) \\
    & torch.reshape(-1, depth)\\
    & maxpool2d(features, kernelsize=1, stride=1)\\
    & torch.nn.Conv1d(inchannels, outchannels, kernelsize)\\
    & torch.cat((features1, features2), dim=1)\\
    \hline
    \multirow{4}{*}{Integration} & features.repeat(1, 1, 1, num copies)\\
  & DepthwiseSeparableConv2d((T+N), D, kernelsize=(T,N), padding=0)\\
  &torch.add(features1, features2)\\
  & nn.Conv2d(kernelsize=3, padding=1, stride=1)\\
    \hline
    \multirow{4}{*}{Successive layers and classifier} & 2 $\times$ [MHSA+nn.Conv2d(kernel size=3, padding=0, stride=1)]\\
    & torch.nn.AvgPool2d(kernelsize=2, stride=2)\\
  &torch.nn.Linear(input, output)\\
  & Softmax\\
  
    \bottomrule
  \end{tabular}
  }
  \caption{\textbf{Architecture of FTP-UniFormerV2-L}.}
  \label{tab:architecture}
\end{table}

\section{Implementation}
\label{sm=sec:model architecture}

\subsection{Model architecture}
\label{sm-sec:model architecture}
We show the architecture details of FTP-UniFormerV2-L/14 in Table~\ref{tab:architecture}. We use the UniformerV2~\cite{li2022uniformerv2} as the video encoder for better and more efficient temporal modeling, and insert the global UniBlocks in the last four layers to perform the multi-stage fusion. For the feature processor, we first reshape the features, and then obtain the spatio-temporal feature through temporal pooling, spatial pooling, and subsequent concatenation. To realize the integration of visual encoder and feature processor, we first replicate the spatio-temporal feature, then use the feature processor to adjust the shape. Features are then summed up element-wise and then passed to the CNN layers to generate the final classification.

\begin{table}[htbp]
  \centering
  \resizebox{0.5\linewidth}{!}{
  \begin{tabular}{cc}
    \toprule
    Parameter & Value\\
    \midrule
    optimizer & AdamW \\
    base learning rate & 1.e-5\\
    weight decay &0.05\\
    optimizer momentum &$\beta1, \beta2$ = 0.9, 0.95\\
    batch size & 512\\
    learning rate schedule &cosine decay\\
    warmup epoch &5\\
    epoch &800\\
    repeated augmentation& 1\\
    flip augmentation &no\\
    augmentation &MultiScaleCrop\\
    \bottomrule
  \end{tabular}
  }
  \caption{\textbf{Parameter settings for training stage 1}: video-text alignment process.}
  \label{tab:setting of step1}
\end{table}

\subsection{Training details}
\label{implementation}

\textbf{Training stage 1: Video-text alignment}. We prompt GPT-4 to generate textual descriptions when provided with a concatenated keyframe image and a specific prompt. The VLM and text encoder remain frozen, we only fine-tune the parameters of the feature processors. We implement the training process of video-text alignment with 40 epochs on the K710 dataset with 16 NVIDIA 80G-A100 GPUs. We adopt repeated augmentation to reduce the video loading overhead. The parameter settings are summarized in Table~\ref{tab:setting of step1}.

\begin{table}[htbp]
  \centering
  \resizebox{0.5\linewidth}{!}{
  \begin{tabular}{cc}
    \toprule
    Parameter & Value\\
    \midrule
    optimizer & AdamW \\
    base learning rate & 1.e-5\\
    weight decay &0.05\\
    optimizer momentum &$\beta1, \beta2$ = 0.9, 0.999\\
    batch size & 128\\
    learning rate schedule &cosine decay\\
    warmup epoch &5\\
    epoch &40\\
    $\alpha$ &0.7\\
    repeated augmentation& 2\\
    label smoothing &0.1\\
    mixup & 0.8\\
    flip augmentation &yes\\
    augmentation &MultiScaleCrop\\
    \bottomrule
  \end{tabular}
  }
  \caption{\textbf{Parameter settings for training stage 2}: model fine-tuning.}
  \label{tab:setting of step2}
\end{table}

\textbf{Training stage 2: Model fine-tuning}
We fine-tune the model trained in stage 1 to adjust the weights so that the model can fit in the integrated features from feature processor and visual encoder. The parameters of the fine-tuning process are shown in Table~\ref{tab:setting of step2}.

\section{Additional Results on Kinetics}
\label{sm-sec:kinetics}
We first report the qualitative results on Kinetics-600, and present additional insights into the improvements on Kinetics-400 when using our FTP framework.

\begin{table}[htb]
  \centering
  \resizebox{\linewidth}{!}{
  \begin{tabular}{lccccc}
    \toprule
    Methods &Input & Crops & TFLOPs & Top-1 & Top-5  \\
    \midrule
   SlowFast R101-NL~\cite{feichtenhofer2019slowfast} &$ 64\times 224^2$ &10× 3 &7.02& 81.8 &95.1 \\
    TimeSformer-L~\cite{bertasius2021space} &$ 96\times 224^2$   & 10 × 3 &7.14& 82.2& 95.6   \\
    ViViT-L FE~\cite{arnab2021vivit} &$ 32\times 224^2$&1 × 3&11.94 & 82.9 &94.6 \\
    MTV-B~\cite{yan2022multiview} &$ 32\times 320^2$ &4 × 3 &4.79 & 83.6 &96.1 \\
    MViT-B~\cite{fan2021multiscale}&$ 32\times 224^2$  &3 × 3 &4.10 & 83.8 &96.3 \\
    MViTv2-B~\cite{li2022mvitv2}&$ 32\times 312^2$     &5 × 3 &1.03 & 85.5 &97.2 \\
    MaskFeat~\cite{wei2022masked}&$ 64\times 224^2$    &4 × 3 &3.77 & 86.4 &97.4\\
    MViTv2-L~\cite{li2022mvitv2}&$ 40\times 352^2$     &5 × 3 &45.48 & 87.9 &97.9 \\
    MaskFeat~\cite{wei2022masked}&$ 40\times 312^2$    &4 × 3 &33.94 & 88.3 &98.0 \\
    VideoMAE V2-H~\cite{wang2023videomae}&$ 16\times 224^2$ & 2 × 3 &17.88 & 88.3& 98.1\\
    VideoMAE V2-g~\cite{wang2023videomae}&$ 64\times 266^2$ & 2 × 3 &38.16 &88.8 &98.2\\
    \midrule
    UniFormerV2-B/16&$ 8\times 224^2$ &4×3 &1.8 &87.4 &97.9\\
    UniFormerV2-L/14 &$ 32\times 224^2$ &2×3 &16.0 &89.5 &98.3  \\
    \midrule
    FTP-UniFormerV2-B/16 (Ours) &$ 8\times 224^2$ &4×3 &2.2 &92.2 &98.9  \\
    FTP-UniFormerV2-B/16 (Ours) &$ 8\times 280^2$ &4×3 &8.1  &93.0 &99.0  \\
    FTP-UniFormerV2-L/14 (Ours)&$ 32\times 224^2$ &2×3 &18.4 &93.8 &99.4  \\
    \textbf{FTP-UniFormerV2-L/14 (Ours)} &$ 32\times 336^2$&2×3 &78.7 &\textbf{94.4} &\textbf{99.3}  \\
    \bottomrule
  \end{tabular}
  }
  \caption{\textbf{Performance on K600}. Best results in \textbf{bold}. We report crops (temporal $\times$ spatial) and TFLOPs for inference.}
  \label{tab:results on K600}
\end{table}

\subsection{Results on Kinetics-600}
We report comparisons on Kinetics-600 with several FTP-UniformerV2 models in Table~\ref{tab:results on K600}. Compared to other models including UniFormerV2, VideoMAE V2, our approach shows consistent improvement. For example, compared to UniFormerV2-B/16, our approach brings 4.8\% improvement. Moreover, when increasing the input size, our model further improves the recognition results. For example, when increasing the spatial input size from $224^2$ to $336^2$, our FTP-UniFormerV2-L/14 shows 0.6\% improvement.

\begin{figure}[htb]
  \centering
  \includegraphics[width=1.0\linewidth]{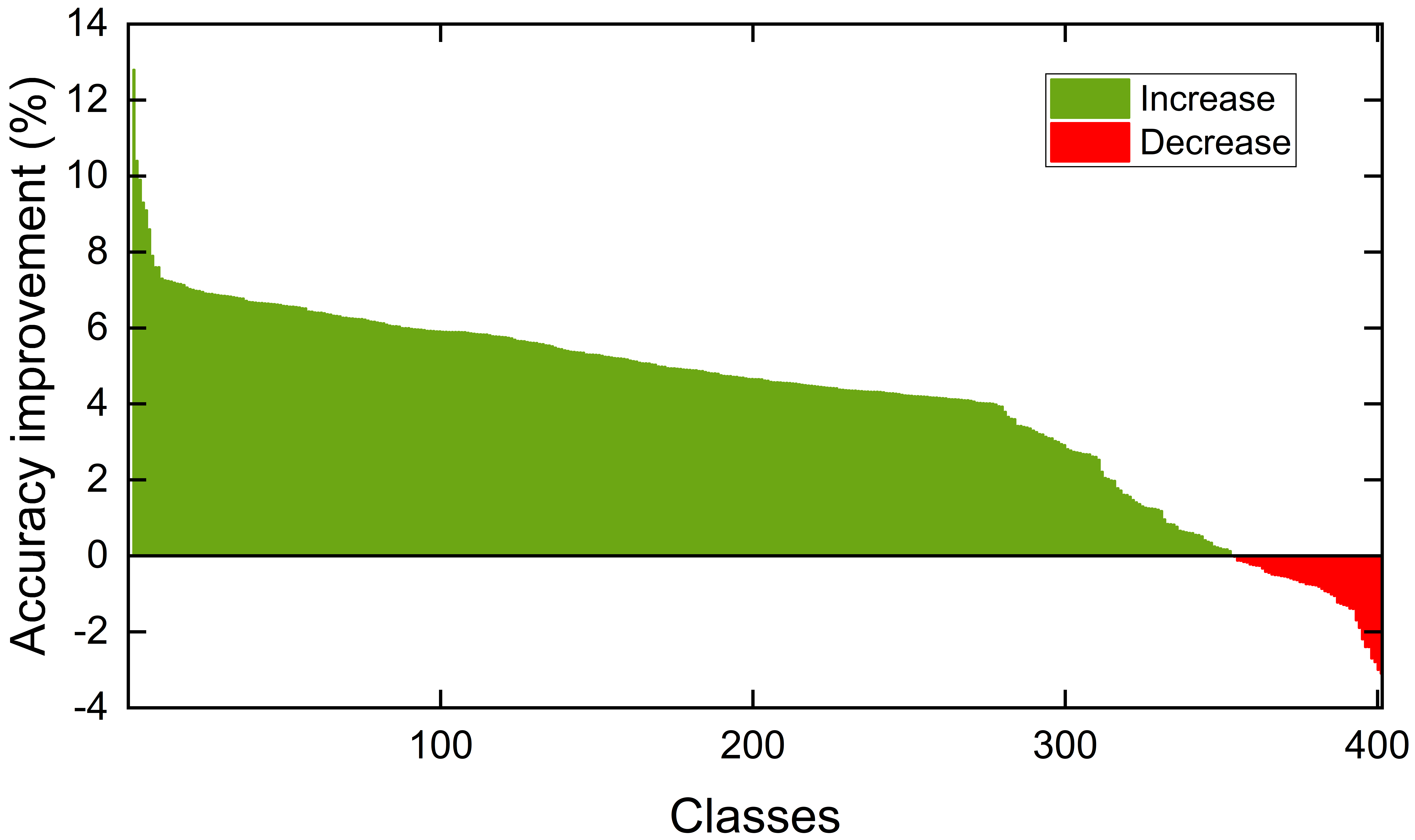}
   \caption{\textbf{Relative accuracy per class on Kinetics-400} by comparing FTP-UniFormerV2-L/14 to a baseline UniFormerV2-L/14 model. The classes are sorted by relative performance.}
   \label{fig:accuracy improvement on all classes}
\end{figure}

\subsection{Class accuracy improvement on Kinetics-400}
We now investigate the relative class performance when using our FTP-UniForm-er
V2-L/14 model, compared to a UniFormerV2-L/14 baseline. We show the relative performance for all classes of Kinetics-400 in Figure~\ref{fig:accuracy improvement on all classes}. For over 85\% of the classes, our method shows improvement. Although there is a lower performance for certain classes, the decrease is typically limited. We further analyze these categories in the following section and investigate the possible reasons for the performance decline.

\begin{figure}[htbp]
  \centering
  \includegraphics[width=1.0\linewidth]{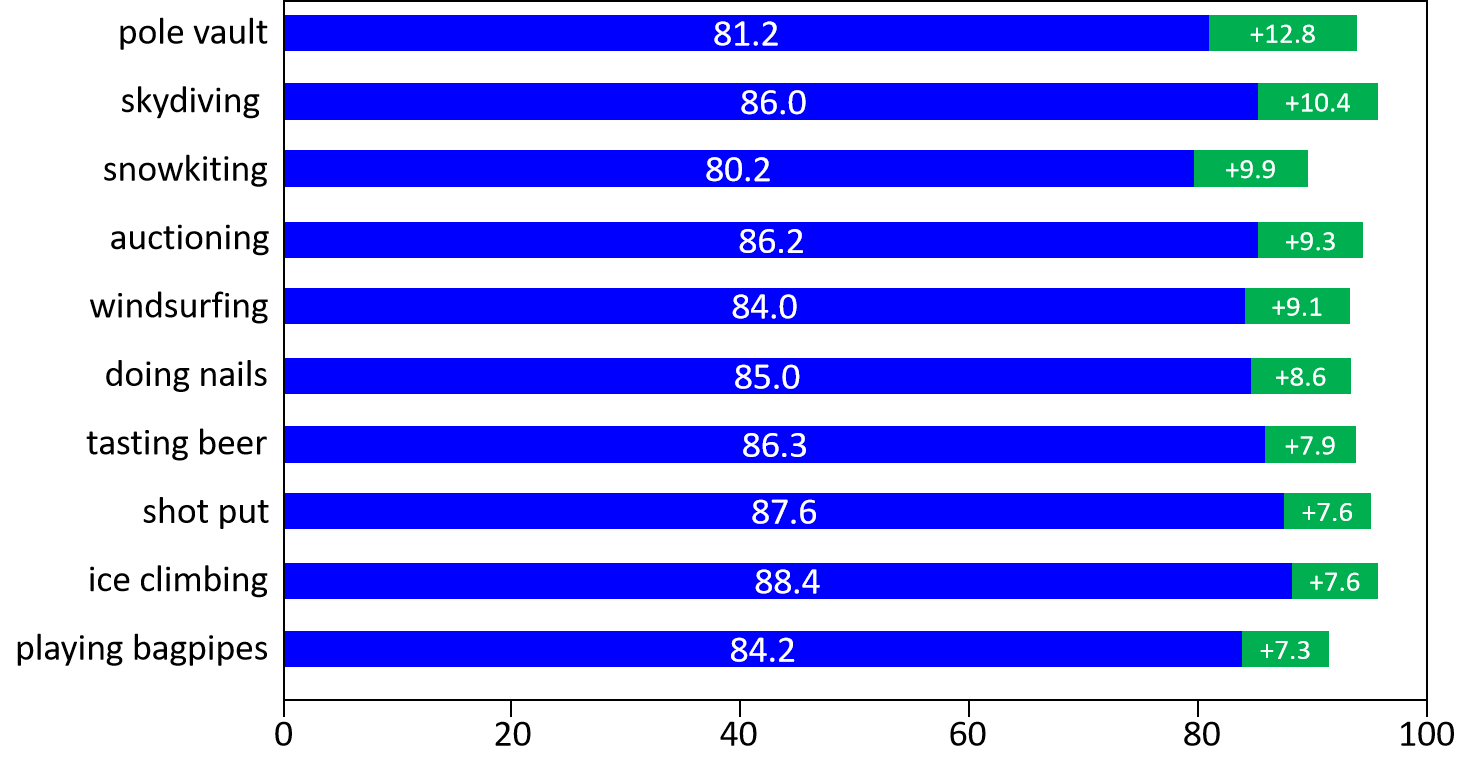}
   \caption{\textbf{Top-10 Kinetics-400 classes with most improvement} when using the FTP framework.}
   \vspace{-10mm}
   \label{sm-fig:top-10_relative_improvement}
\end{figure}

\begin{figure}[htbp]
  \centering
  \includegraphics[width=1.0\linewidth]{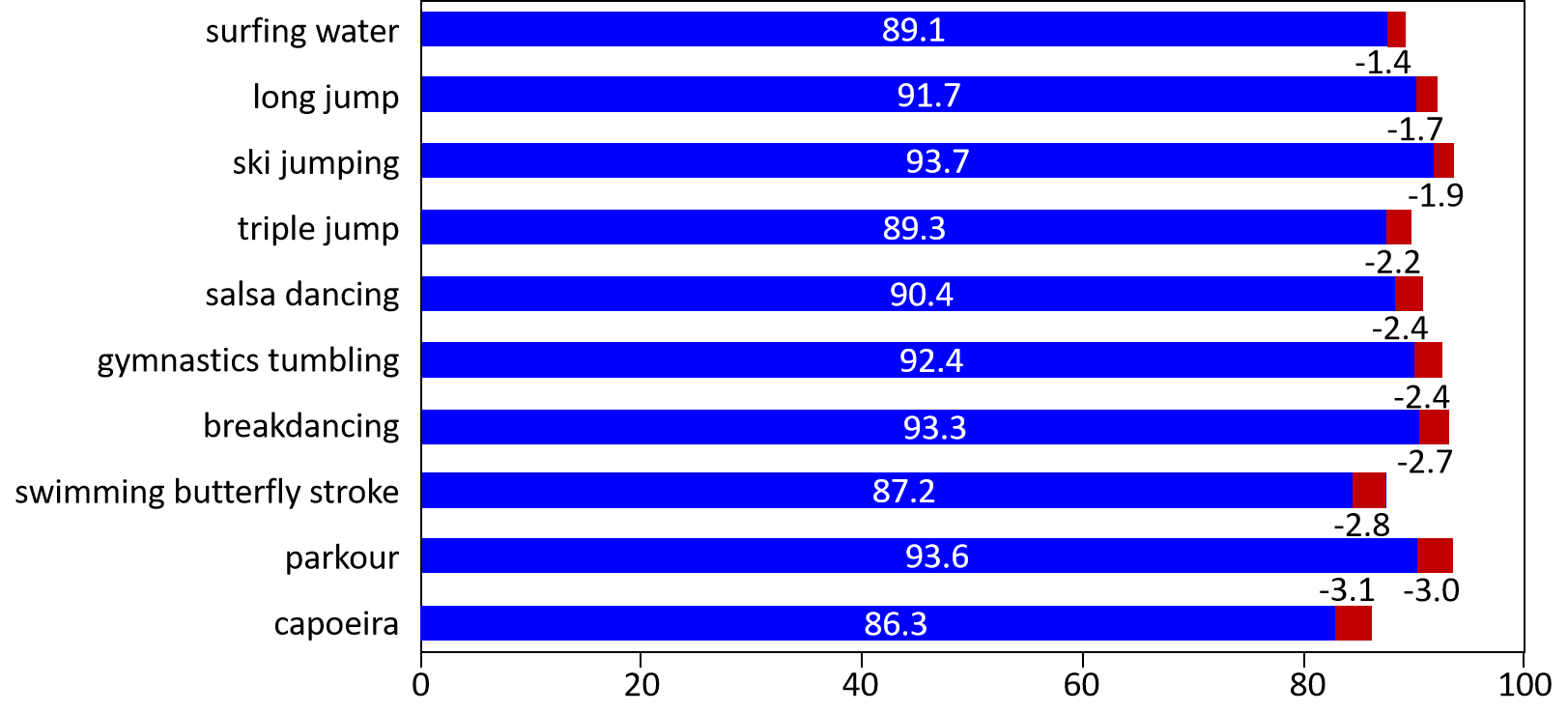}
   \caption{\textbf{Top 10 Kinetics-400 classes with least improvement} when using the FTP framework. These classes have a lower score when using our FTP framework.}
   \label{sm-fig:bottom-10_relative_improvement}
\end{figure}

\textbf{Top-10 classes with highest improvement}. We show the top-10 classes from Kinetics-400 that have seen the most significant improvement when using our FTP framework in Figure~\ref{sm-fig:top-10_relative_improvement}. These classes correspond to the leftmost 10 bars in Figure~\ref{fig:accuracy improvement on all classes}.

These 10 classes are all challenging and require the integration of action objects, context, and other information. For instance, the ``auctioning" class typically involves identifying the context of objects and actions within the video, and our method yields a +9.3\% improvement compared to UniformerV2.

\textbf{Top-10 classes with biggest performance decline}. Application of our method performs worse than a baseline Uniformerv2 model on a minority of the classes of Kinetics-400. We analyze the bottom-10 classes in terms of relative improvement in Figure~\ref{sm-fig:bottom-10_relative_improvement}. We see several action in which the scene or specific recording setting might be introducing distracting elements into the textual representations. For example, long jump and triple jump take place in stadiums with many sports occurring simultaneously. Parkour, capoeira, and breakdancing can be performed in scenes that vary significantly.

Because the VLM is presented only with several key frames, motion information is not emphasized. As a result, the motion information that is initially present in the visual encoder might be diluted in the presence of the textual encodings. This might have deteriorated the performance for classes such as salsa dancing and gymnastics tumbling.

\textbf{Visualization of embeddings}. To further analyze the effect of our prompts, we use t-SNE~\cite{van2008visualizing} and visualize the distribution of features from different classes. We use the 2048D feature activations of the first layer of the classifier.

\begin{figure}[htbp]
  \centering
  \includegraphics[width=1.0\linewidth]{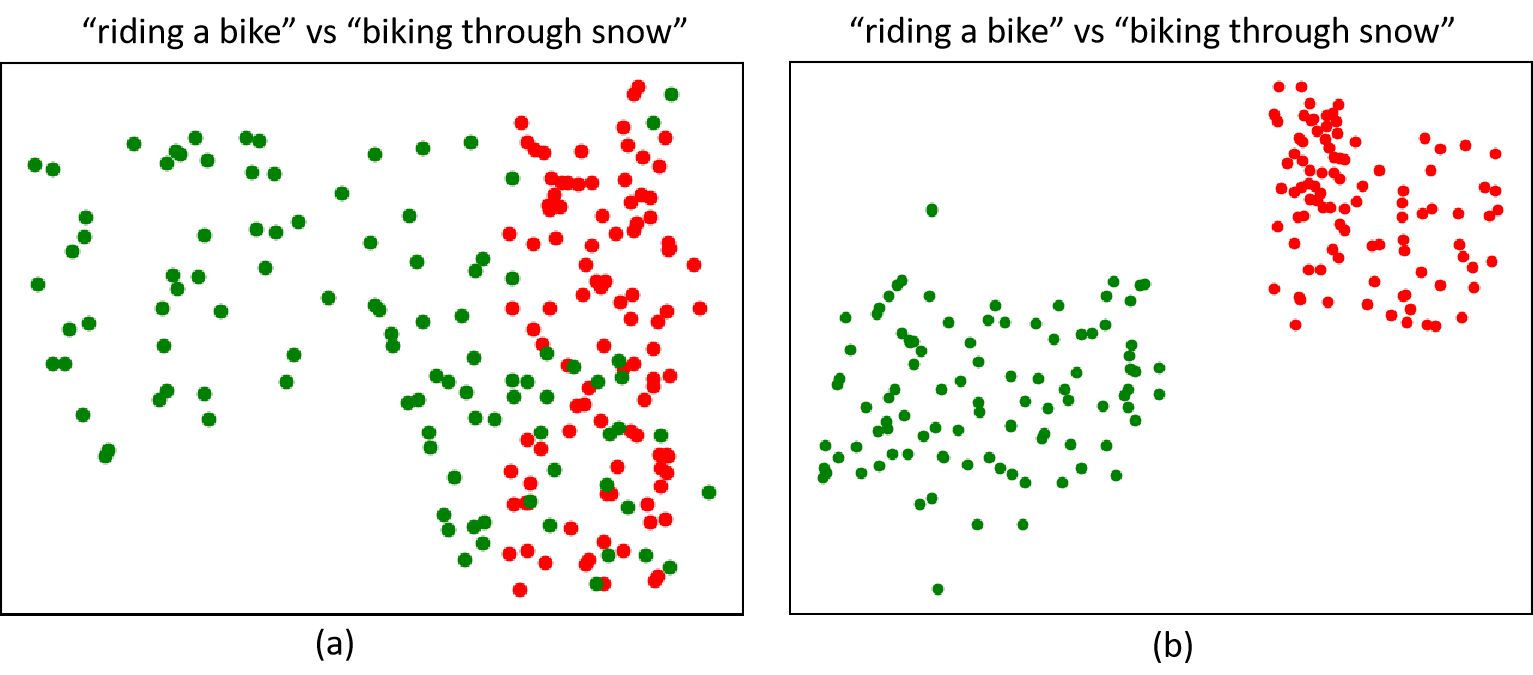}
   \caption{\textbf{t-SNE visualization of feature distribution} of “riding a bike” vs “biking through snow” (a) without applying the FTP framework, and (b) when applying FTP.}
   \label{sm-fig:tsne-example}
\end{figure}

In Figure~\ref{sm-fig:tsne-example}, we randomly select 100 samples from the ``riding a bike" and ``biking through snow" category. The distribution of features from each class is much more compact and discriminative after applying our FTP approach.

\begin{figure}[htbp]
  \centering
  \includegraphics[width=1.0\linewidth]{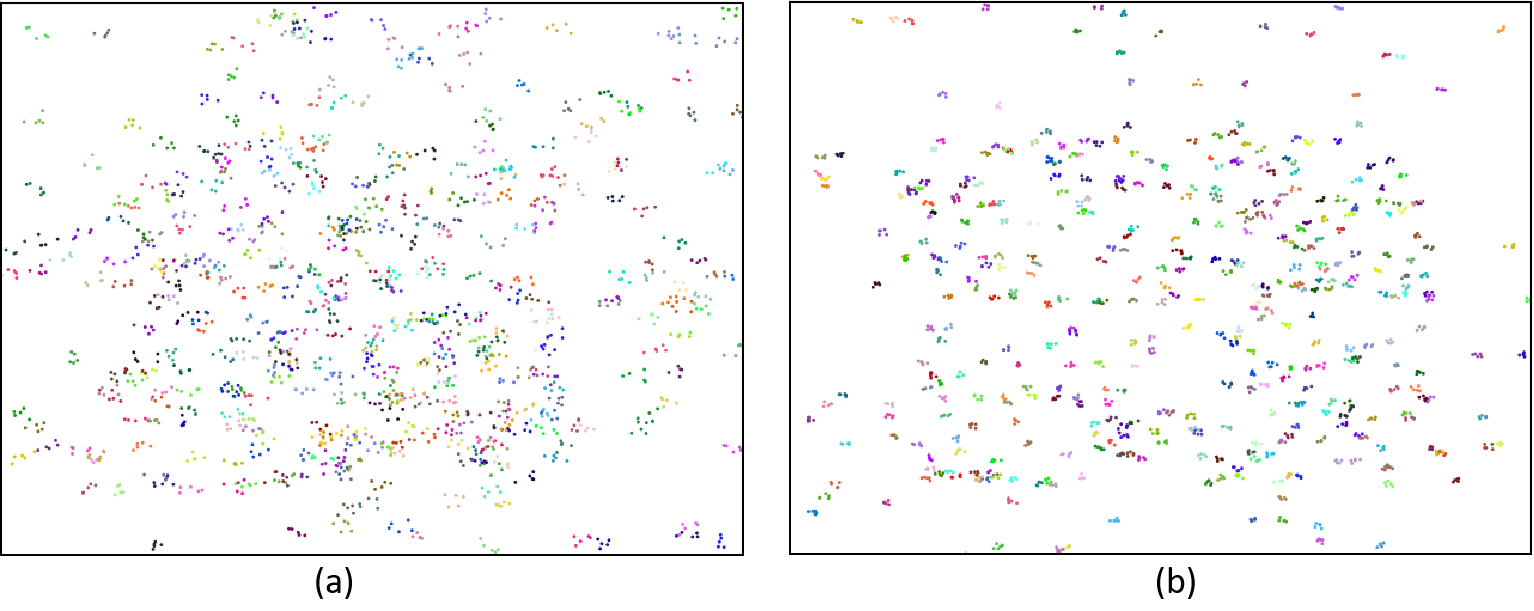}
   \caption{\textbf{t-SNE visualization of feature distribution of all Kinetics-400 classes} (a) without applying the FTP framework, and (b) when applying FTP.}
   \label{sm-fig:tsne-all}
\end{figure}

In Figure~\ref{sm-fig:tsne-all}, we randomly select 5 samples of each of the 400 categories of Kinetics-400. After applying our FTP approach, we observe that the intra-class variation of features becomes smaller, while the inter-class variation increases. This indicates that the classes are better isolated, consequently benefitting classification.

\section{Effect of Number of Keyframes $K$ to VLM}
\label{sm-sec:ablation_keyframes}
In our experiments in the main paper, we used $K = 5$ as default to prompt the VLM. To evaluate the effect of using less or more keyframes, we conducted experiments using FTP-UniFormerV2-L/14 with different values of $K$. We kept the experiment settings consistent with those used in the previous experiments on Kinetics-400 and summarize the results in Table~\ref{tab:number of keyframes}.

We can observe that when $K = 1$, the performance is the worst, with a top-1 accuracy of 91.9\%. As $K$ increases to 3, the top-1 accuracy improves to 92.5\%. The best performance of 93.4\% is achieved for $K = 5$. However, as we continue to increase $K$, the performance starts to decline, reaching 93.2\% for $K = 7$. This could be because, at this point, the image descriptions generated by GPT become overly complex, and potentially focus on less consistent aspects of the video. Consequently, this might introduce some noise in the text encodings that affects the performance.

\begin{table}[htbp]
\centering
\begin{tabular}{lcccc}
\toprule
                   & $K = 1$   & $K = 3$   & $K = 5$   & $K = 7$   \\
Top-1 accuracy &  91.9   & 92.5    & \textbf{93.4}    &  93.2 \\
\bottomrule
\end{tabular}
\caption{\textbf{Effect of number of concatenated keyframes $K$} on the performance on Kinetics-400 (in \%).}
\label{tab:number of keyframes}
\end{table}

\end{appendices}

\end{document}